\documentclass{Interspeech}
\usepackage{algorithm}
\usepackage{algorithmic}
\usepackage{amsmath}

\interspeechcameraready

\title{Weight Factorization and Centralization for Continual Learning \newline in Speech Recognition}

\author[affiliation={1,*}]{Enes Yavuz}{Ugan}
\author[affiliation={2,*}]{Ngoc-Quan}{Pham}
\author[affiliation={1,2}]{Alexander}{Waibel}

\affiliation{Interactive Systems Lab}{ Karlsruhe Institut of Technology (KIT)}{Germany}
\affiliation{InterACT}{ Carnegie Mellon University (CMU)}{USA}
\email{enes.ugan@kit.edu, ngocquap@andrew.cmu.edu, alex@waibel.com}

\usepackage{comment}

\begin{document}

\maketitle
\begingroup
\renewcommand\thefootnote{*}   
\footnotetext{Equal contribution}
\endgroup
\begin{abstract}
Modern neural network based speech recognition models are required to continually absorb new data without re-training the whole system, especially in downstream applications using foundation models, having no access to the original training data. Continually training the models in a rehearsal-free, multilingual, and language agnostic condition, likely leads to catastrophic forgetting, when a seemingly insignificant disruption to the weights can destructively harm the quality of the models. Inspired by the ability of human brains to learn and consolidate knowledge through the waking-sleeping cycle, we propose a continual learning approach with two distinct phases: factorization and centralization, learning and merging knowledge accordingly. Our experiments on a sequence of varied code-switching datasets showed that the centralization stage can effectively prevent catastrophic forgetting by accumulating the knowledge in multiple scattering low-rank adapters. 

\end{abstract}

\section{Introduction}



Large-scale multilingual speech models~\cite{radford2023robust} have become the backbone of speech applications, thanks to the high performance achieved in multiple languages, enabled by scaling the models, computation and data. Despite such achievement, the one-size-fit-all solution might still require further fine-tuning, in order to optimize the performance on certain tasks or datasets, for example dealing with code-switching~\cite{huber2022code,huang2003automatic} or low-resourced languages~\cite{suhm1994towards,stuker2003multilingual}.
Dealing with such data is crucial in many systems such as simultaneous speech translation systems~\cite{waibel2012simultaneous,waibe112005chil}. There is a variety of methods that allow for fast and efficient adaptation, such as using low-dimensional adapters~\cite{pham2021efficient,hu2021lora}, but this has often been treated as a one-off solution. More appropriately, we can treat the models as \textit{continual learning} agents, in a scenario that these agents continually learn to adapt to the datasets without losing the previously learned knowledge. In fact, it only takes a few updates on a new dataset with a completely different distribution than the training data (which is not accessible when it comes to foundational models) to cause catastrophic forgetting, when the improvement on the downstream task is compromised with the rapid deterioration of the original model. 

The continual learning research area currently addresses this challenge primarily through three approaches: \textit{rehearsal}-based techniques rely on a portion of the training data to partially recover performance loss \cite{robins1995catastrophic, NIPS2017_0efbe980}; \textit{regularization} techniques penalize gradients or weights from deviating from their original state \cite{kirkpatrick2017overcoming}; and \textit{architecture-based} techniques expand the model capacity \cite{zhang2020regularize, razdaibiedina2023progressive} to accommodate new data—or a combination of these approaches altogether~\cite{pham23_interspeech}. It is imperative to find solutions for continual learning on top of foundational models, especially when access to data is limited or unavailable, which makes many techniques difficult to apply.

On the other hand, having a strong foundation also enables other possibilities compared to approaches that rely on training models from scratch. First of all, the foundational models can be adapted on new small datasets very rapidly with low-rank adapters, which are additional weights assigned to specific datasets/tasks. These weights are invoked only when these tasks are identified or given during inference. Consequently, the incoming dataset that is given to the model can be scattered into segments which are assigned to separated adapters. This observation alleviates the problem of \textit{learning order} in continual learning formulations, in which the order of the input datasets can affect the final performance, because the later datasets potentially harm the performance of the previous ones. Thirdly, linear combination of low-rank adapters has been used to make the model adapt to several conditions at once, but using this practice to alleviate catastrophic forgetting in the base model has not been studied.

Motivated by the works using stochastic weight averaging~\cite{izmailov2018averaging} to obtain robost models against catastrophic forgetting, as well as federated learning techniques combining the information from multiple scattered model copies \cite{douillard2025streaming}, in this manuscript we propose a factorization-centralization framework to enable rehearsal-free continual learning with foundational models. Here, our key idea is to divide the learning process into two stages. In the factorization stage, the incoming data stream containing multiple datasets is scattered and allocated in different low-rank adapters, effectively expanding the model capacity in learning. Naturally, without any powerful regularization all of these adapters cause detrimental effects on the overall performance (except for the downstream segments that they are trained for). The centralization in turn merges the adapters in order to recover such performance. Motivated by the previous works showing that averaging fine-tuned adapters can improve accuracy, we empirically found that averaging can also mitigate the catastrophic forgetting effect, with an intuition of relying on a simple Gaussian Prior for the weights. As a result, this is a simple and scalable approach to continual learning with large foundational models.  

Our experiments focus on 6 code-switched datasets, chosen as study targets because one of the current state-of-the-art models, Whisper, cannot recognize them effectively. With the goal of obtaining a continually learned model that can balance between stability - keeping the previous knowledge, and plasticity - learning to adapt to new conditions, we found that this strategy provides the best trade-off, compared with the sequential learning approach using regularization via stochastic weight averaging and distillation. Another advantage comes from the ease to apply, with the only hyper-parameter to select being the size of each iteration before centralization.

\section{Related Works}
\label{sec:cl}

We consider a setting where an automatic speech recognizer (ASR) is trained by observing the datasets [$D_1, D_2, \dots D_T$] sequentially. Each dataset $D_t$ consists of samples $(x_t^i, y_t^i)$ being the input utterances and labels respectively. At each iteration $t$, not only does the model parameters $\theta_t$ have to optimize for the current dataset $D_t$ starting from $\theta_{t-1}$, but it also has to minimize the empirical risk on all of the seen datasets $D_{<t} = D_{1\dots t-1}$ so far, with the cumulative risk.

\begin{equation}
    \label{eq:cumulativerisk}
    L_t = \sum_{i}^{N_t}-\log p(y_t^i |x_t^i, \theta_t) + \sum_{j=1}^{t-1}\sum_{k}^{N_j}-\log p(y_t^k |x_t^k, \theta_t)
\end{equation}

This equation is in general difficult to achieve, because generally the datasets from $D_{<t}$ are not accessible after their training turns. Furthermore, sequentially optimizing $\theta$ for each dataset is less optimal than having all datasets at once. Since the weight updates of $D_t$ can conflict with previous datasets, training on $D_t$ can easily deteriorate the quality for $D_{<t}$. This particular problem of catastrophic forgetting has been shown in multilingual speech recognition~\cite{pham23_interspeech}, happening with even large scale foundation models~\cite{baevski2020wav2vec,radford2023robust}.

In practice, Equation~\ref{eq:cumulativerisk} is realized via the combination of the current loss function $L_t$ (left hand side) combined with a regularization term $R$ preventing the model to conflict with $D_{<t}$. Access to previously trained data (even partially) is important, enabling regularization based on data-driven weight importance~\cite{kirkpatrick2017overcoming,zenke2017continual} or variational estimation of distributional parameters in Bayesian methods~\cite{nguyen2017variational,farquhar2019unifying}, or simply rehearsing and recovering the lost performance~\cite{lopez2017gradient}. When such data is not available, using expansion-based techniques such as adapters~\cite{pham2021efficient,pham23_interspeech,hu2021lora} or learnable prompts~\cite{smith2023coda} to create an independent processing path in the models, being enabled only for the task $D_t$, is common practice. While this has been proven to be useful, especially in multilingual~\cite{qian24_interspeech,pham23_interspeech} or speaker-based adaptation~\cite{huber2024continuously}, the downside is to rely on task boundaries and defeats the original purpose of powerful language agnostic speech recognizers. Moreover, task boundaries between speech datasets might be blurry, and multilingual speech models are often required to be language and accent agnostic for actual applications. Given these contraints, stochastic weight averaging~\cite{izmailov2018averaging} might be a promising approach, when the current model state is ensured to be stable, via an exponentially moving average of the weights before and after updating.


\section{Main approach}
\label{sec:approach}

Motivated by the daytime–nighttime cycle and the role of sleep in memory consolidation in the human brain, we propose decoupling the typical learning and regularization objectives in continual learning into two distinct phases. In the \textit{factorization} phase, we aim to expand the knowledge base model using temporary, learnable adapters on the datasets to be learned. In turn, the \textit{centralization} phase compresses the learned elements in the adapters back into the knowledge base. While the first phase—using the pretrained model as the knowledge base and adding adapters for specific tasks and datasets—is commonly used, the second phase is challenging because the compression must be performed without access to the original data used to train the foundation model.

\begin{figure*}[t!]
   \centering
  \includegraphics[scale=0.3]{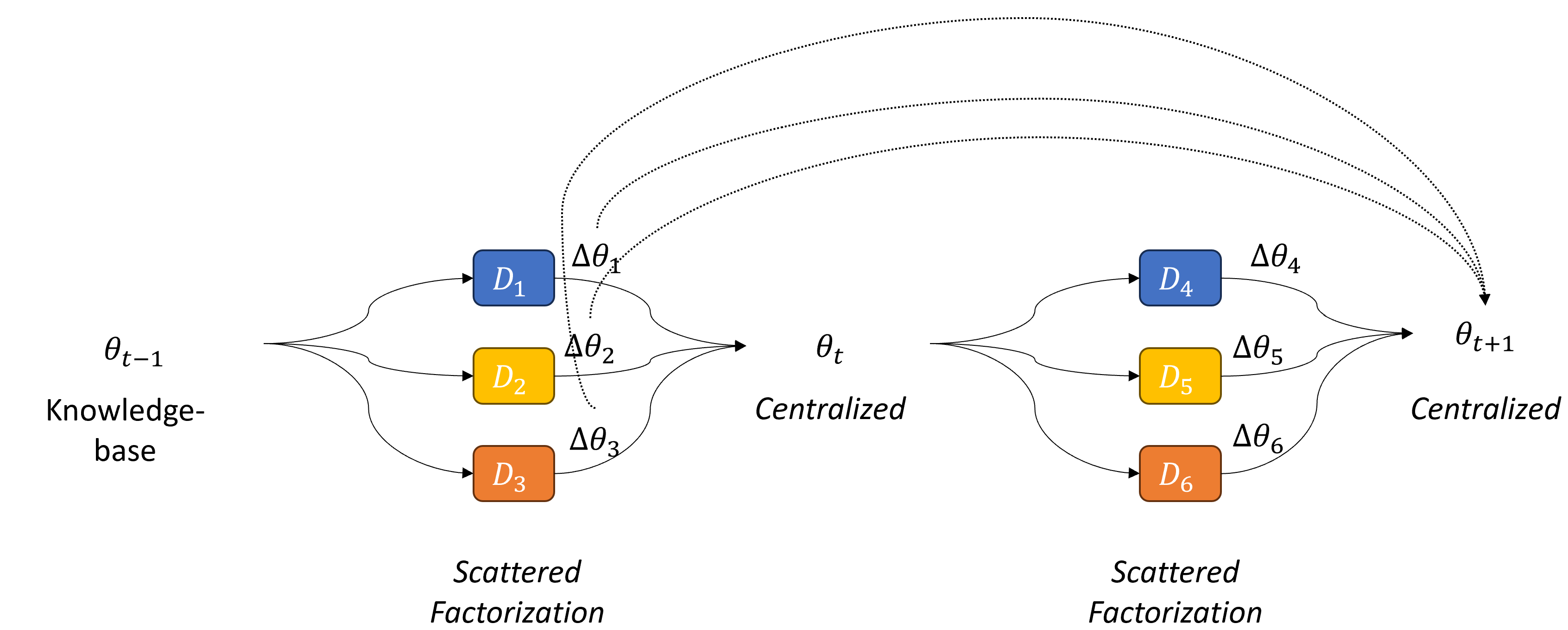}

 \caption{Continual learning with factorization and centralization: Data streams are temporally segmented, each with dedicated low-rank weights. Periodically, these weights (e.g., from 3 language-type datasets per iteration) are averaged and centralized into the base model.}
  \label{fig:freevc-arch}
\vspace{-5mm}
\end{figure*}
\vspace{-0.1cm}

\subsection{Factorization Stage}

In our framework, the model is continually updated with an incoming stream of datasets. Instead of assigning a single adapter to the entire stream, we distribute the incoming knowledge across multiple adapters. When clear data boundaries exist---for example, with separately collected datasets---we assign a dedicated adapter for each dataset. In scenarios where such boundaries are not explicit, its possible to allocate adapters based on capacity (e.g., each adapter is responsible for a fixed number of samples). This strategy helps to mitigate the influence of dataset ordering on the learning process (prior to each centralization step).

To efficiently integrate new information, we leverage low-rank adapters (LoRA). Each adapter introduces only a small number of additional parameters while capitalizing on the rich features provided by the pretrained knowledge base. Concretely, consider a linear layer with an original projection matrix $\mathbf{W}$ and an input feature $\mathbf{x}$. Rather than modifying $\mathbf{W}$ directly, we create a parallel path with a low-rank update $\Delta \mathbf{W}$ defined as:
\begin{equation}
\Delta \mathbf{W} = \frac{\alpha}{r}\,\mathbf{A}\mathbf{B},
\end{equation}
where $\mathbf{A} \in \mathbb{R}^{d_{\text{out}} \times r}$ and $\mathbf{B} \in \mathbb{R}^{r \times d_{\text{in}}}$ are learnable matrices, $r$ is the rank (with $r \ll \min(d_{\text{in}}, d_{\text{out}})$), and $\alpha$ is a scaling factor controlling the magnitude of the update.

Thus, the output of the adapted linear layer becomes:
\begin{equation}
\mathbf{y} = \mathbf{W}\mathbf{x} + \Delta \mathbf{W}\mathbf{x} = \mathbf{W}\mathbf{x} + \frac{\alpha}{r}\,\mathbf{A}\left(\mathbf{B}\mathbf{x}\right).
\end{equation}

This formulation enables each adapter to efficiently learn new information from its allocated subset of data while keeping the overall parameter overhead minimal. In practice, these weights are added to the query and key linear layers in the Transformer models.

\subsection{Centralization Stage}

Given the scattered low-rank weights learned for each dataset, how can we proceed to the centralization stage? In our preliminary experiments, adding the adapters individually can only improve the performance of the code-switched task at hand (for example Arabic-English), while causing catastrophic forgetting, reflected in the rapid deterioration in other languages (such as German). 

The centralization process is depicted in Algorithm~\ref{alg:continual_learning}. The knowledge-base model is updated periodically after $K$ datasets, each of which are trained separated using adapters. When centralization happens (line 7), we accumulate all of the adapters trained up to the current time steps via averaging (line 8), and merge them into the base model (line 9). 

Even though the algorithm suggests that we have to store all of the low-rank adapters trained from the beginning, line 8 and 9 can be efficiently implemented by storing the $\Delta_{\text{avg}}$ after each centralization, which is rescaled and added to the new adapters $\Delta\theta$ in the next datasets. The total extra memory required is $K \times $ adapter memory. 

\begin{algorithm}
\caption{Continual Learning with Waking and Sleeping}
\label{alg:continual_learning}
\scriptsize
\begin{algorithmic}[1]
\REQUIRE Datasets $D_1, D_2, \dots, D_N$, number of adapters per centralization $K$, initial model weights $\theta_0$ - Knowledge Base (KB)
\FOR{$t = 1$ \TO $N$}
    \STATE Create adapter $\Delta\theta_t$.
    \FOR{$i = 1$ \TO $|D_t|$}
        \STATE Compute loss: $\mathcal{L}_t^i = -\log p(y_t^i \mid x_t^i, \theta_{t-1}, \Delta\theta_t) $.
        \STATE Update $\Delta\theta_t$ using Stochastic Gradient Descent.
    \ENDFOR
    \IF{$t \bmod K == 0$}
        \STATE Merge adapters: $\Delta_{\text{avg}} = \frac{1}{t} \sum_{j=1}^{t} \Delta\theta_j$.
        \STATE Update KB: $\theta_t = \texttt{lora\_merge}(\theta_{t-1}, \Delta_{\text{avg}})$.
    \ELSE
        \STATE Set $\theta_t = \theta_{t-1}$.
    \ENDIF
\ENDFOR
\end{algorithmic}
\end{algorithm}

Intuitively we would like $\Delta_{\text{avg}}$ being close to 0, or at least being sparse, to avoid disruption to the knowledge base $\theta_0$. Therefore, the low-rank weights are regularized with weight decay, which is based on the Bayesian interpretation imposing a Gaussian Prior on the weight with mean 0. Without the pre-training data, it is not possible to use data-driven algorithms to calculate the per-weight elasticity~\cite{kirkpatrick2017overcoming}, but we do expect an average of Gaussian distributions to yield smaller variance and potentially becoming sparser the more low-rank adapters being averaged. The mathematical justification is as follows:


Suppose we have $N$ independent models, each with weights drawn from a Gaussian distribution:

\[
w_i \sim \mathcal{N}(0, \sigma^2)
\]

where $w_i$ represents the weights of the $i$-th model.

Now, define the averaged weights:
\[
\bar{w} = \frac{1}{N} \sum_{i=1}^{N} w_i
\]
Since each $w_i$ is independently drawn from $\mathcal{N}(0, \sigma^2)$, the expectation of $\bar{w}$ is:
\vspace{-8pt}
\[
\mathbb{E}[\bar{w}] = \mathbb{E} \left[\frac{1}{N} \sum_{i=1}^{N} w_i \right]
= \frac{1}{N} \sum_{i=1}^{N} \mathbb{E}[w_i]
= \frac{1}{N} \sum_{i=1}^{N} 0 = 0
\]
Thus, the mean remains zero. Since $w_i$ are independent, the variance of $\bar{w}$ is:

\[
\text{Var}(\bar{w}) = \text{Var} \left( \frac{1}{N} \sum_{i=1}^{N} w_i \right)
\]

Using the property that the variance of a sum of independent variables is the sum of their variances:
\[
\text{Var}(\bar{w}) = \frac{1}{N^2} \sum_{i=1}^{N} \text{Var}(w_i)
= \frac{N \sigma^2}{N^2} = \frac{\sigma^2}{N}
\]
\vspace{-11.5pt}


Thus, as we average more models, the standard deviation decreases proportionally to $1/\sqrt{N}$, meaning that the distribution of the averaged weights becomes more concentrated around 0. This equation suggests that, the more factorization we have before centralization, the more possible it is to be robust to forgetting (when the summation of knowledge-base and adapters is less likely to deviate too far). In practice, however, model weights have different standard deviation because the regularization term is dominated by the main cross entropy loss, which depends on the data. 

It is notable that, the adapters learned after centralization are conditioned on the previously learned adapters, being different than the common practice that the base model remains the same indefinitely while increasing the number of adapters per dataset/tasks. 

\vspace{-8pt}

 \begin{table}[h]
     \caption{Backward evaluation of baseline and adapted models. For Mandarin, the evaluation metric is CER, all other data sets are evaluated using WER. }
     \small
     \setlength\tabcolsep{2px}
     \renewcommand{\arraystretch}{1}
     \centering
     \begin{tabular}{|c|c|c|c|c|c|c|} \hline
        Model  & English & German & Arabic & Turkish & Mandarin &Spanish \\
     \cline{2-7}
     & TED & MLS & MGB  & TSC & AISHELL & CV \\
     \hline
     \hline
     Base & 6.0 & 6.0 & 23.5 & 16.7 & 9.3 & 6.1 \\
     \hline
     \hline
     ArZen   & 5.8 & 10.1 & 49.0 & 43.5 & 83.6 & 29.0  \\
     SEAME  & 7.9 & 23.4 & 132.4 & 151.8 & 34.1 & 53.4  \\
     Fisher & 5.7 & 14.1 & 49.6 & 30.5 & 35.3 & 9.5\\
     \hline
     Centralized   & \textbf{4.3} & 7.7 & 25.4 & 15.4 & 8.5 & 6.8  \\
     \hline
     Ascend & 6.5 & 7.2 & 25.5 & 18.6 & 10.3 & 6.5 \\
     TalCS & 7.0 & 8.9 & 32.7 & 20.7 & 12.2 & 10.1 \\
     Tunswitch & 21.4 & 64.5 & 79.4 & 54.8 & 82.1 & 53.3 \\
     \hline
     2nd Cent. & 4.8 & \textbf{6.0} & \textbf{19.9} & \textbf{14.3} & \textbf{6.9} & \textbf{5.8} \\
     \hline
     SWADT\cite{vander2023rehearsal} & 5 & 6.3 & 24.2 & 16.2 & 8.4 & 6.7 \\
     \hline
     \end{tabular}
     \label{tab:backward}
 \end{table}

 \begin{table*}[t!]
     \caption{Left part: Forward evaluation comparison of baseline models and our newly proposed conttinual learning framwework. Right part: Backward evaluation comparison of baselines with proposed continual learningn framework }
     \small
     \setlength\tabcolsep{2px}
     \renewcommand{\arraystretch}{1}
     \centering
     \begin{tabular}{|c|c|c|c|c|c|c|c||c|c|c|c|c|c|c|c|c|} \hline
        Model  &  Fisher & ArZen & SEAME & TalCS & Ascend & Tunswitch & AVG & DECM & En & De & Ar & Tr & Zh & Es & AVG\\
     \cline{2-16}
     & WER & WER & MER & MER & MER & WER &  & WER & WER & WER & WER & WER & CER & WER & \\
     \hline
     \hline
     Base & 29.4 & 52.8 & 42.4 & 16.9 & 19.6 & 75.3 & 39.4 & 9.8 & 6 & 6 & 23.5 & 16.7 & 9.3 & 6.1 & 11.1\\
     Fine-tuned & 21.4 & 38.1 & 22.1 & 15.7 & 22.6 & 31.8 & 22.7 & 25.3 & 6.9 & 21.4 & 50.5 & 25.4 & 14.5 & 14.6 & 22.7\\
     \hline
     \hline
     SWADT  & 25.7 & \textbf{43.4} & 25.7 & 13.6 & 13.3 & 56.5 & 29.7 & 12.4 & 5 & 6.3 & 24.2 & 16.2 & 8.4 & 6.7 & 11.3\\
     \hline
     2nd Cent. \textbf{(ours)}  & \textbf{23.3} & 45.1 & \textbf{24.5} & \textbf{12.1} & \textbf{12.1} & \textbf{55.3} & \textbf{28.7} &  \textbf{10.7} & \textbf{4.8} & \textbf{6} & \textbf{19.9} & \textbf{14.3} & \textbf{6.9} & \textbf{5.8} & \textbf{9.8}\\
     \hline
     \end{tabular}
     \label{tab:eval-mono}
     \vspace{-3mm}
 \end{table*}

\section{Experiments}

\subsection{Experiment Setup}
Code-switching—rapid alternation of languages within the same utterance—remains harder than monolingual ASR because switch points are brief, pronunciation patterns overlap, and the available speech often exhibits distributional shifts such as telephone speech and data sparsity.
Early work applied language identification \cite{schultz1996lvcsr} which would then be used for monolingual decoding; subsequent studies trained multilingual acoustic models that still lagged behind single-language systems \cite{schultz2001experiments}; later, data-augmentation that synthesizes mixed-language speech boosted language agnostic ASR models code-switching abilities \cite{ugan2022language}.
Although the Whisper foundation model~\cite{radford2023robust} delivers strong multilingual ASR, its code-switching accuracy is modest.
In our experiments we therefore focus on adapting Whisper’s \emph{large-v3-turbo} variant.
Forward evaluation uses six code-switching corpora: ArZen~\cite{hamed2020arzen}, Fisher~\cite{weller-etal-2022-end}, SEAME~\cite{lyu2010seame}, TUNSwitch~\cite{abdallah2024leveraging}, ASCEND~\cite{lovenia-etal-2022-ascend}, and TalCS~\cite{li2022talcs}, grouped $K\!=\!3$ at a time.  
Because we care about overall recognition accuracy, fine-grained switching metrics such as PIER~\cite{ugan2025pier} are omitted.  
For backward evaluation we report results on TED \cite{hernandez2018ted} (en), MLS \cite{Pratap2020MLSAL} (de), MGB \cite{ali2016mgb} (ar), TSC \cite{mussakhojayeva2023multilingual} (tr), AISHELL \cite{aishell_2017} (zh), CommonVoice \cite{ardila2019common} (es), and DECM \cite{ugan2024decm} (de–en CS).

For comparison, we followed the continual learning approach that utilizes stochastic weight averaging~\cite{izmailov2018averaging} and distillation~\cite{buzzega2020dark} (also known as Dark Experience Replay) to maintain the stability of the training course, referred as the \textbf{SWADT} model. To the best of our knowledge, this is a scalable approach applicable to end-to-end speech recognition and was demonstrated to outperform even rehearsal-based techniques. 




\subsection{Centralization serendeipitously enables backward transfer}

Among the benchmark testsets that we use to evaluate backward transferring, it is observable that German, modern standard Arabic and Turkish are not covered by the incoming training datasets. They do feature English vocabularies, but spoken in a heavily accented way and most of the data still contain most of the original languages. Therefore, we expect that each factorized adapter should cause catastrophic forgetting towards these languages. As can be seen from Table 1, when the adapters are added via ArZen, Seame or Fisher, they dramatically increase the error rates in the backward benchmark. For example, Arabic and Turkish tests have over 100\% error rates with the SEAME factor. The only scenarios in which the error rates do not change dramatically is English (TEDTalks) with any adapter, or German with the ArZen adapter. Even though its notable that these datasets are relatively small, and each rapid adaptation only needs a few thousands of updates. With a rather unstrict regularization scheme in our centralization, its not expected to observe any mitigation, since weight averaging has often been seen as an alterative to model ensembling.

In contrary to such expectation, we even observed positive backward transfer after centralization. For the first iteration, the centralized model delivered an improvement over the adapters and even the base model for English, Turkish, Mandarin while the other tests have been ``recovered'' to the procimity of the original performance. 

For the second centralization stage, the backward transfer effect is observed even more strongly in all of the test sets except English (when the 2nd iteration is slightly worse than the 1st counterpart). In comparison with the SWADT baseline model, it can also impressively exhibit the backward transfer property for Turkish, Mandarin, English and Spanish, however to the lesser extent compared to our approach. We measured the sparsity of each $\Delta\theta_t$ and noticed that the averaged model has much higher sparsity than any other adapter, except the Ascend one\footnote{Training is early-stopped quickly which might explain the high sparsity of the adapters.}.

\subsection{Balancing between learning and regularization}

Table~\ref{tab:eval-mono} shows the performance of our Centralized model compared to the baseline (SWADT). We used a fine-tuned model on 6 datasets using LoRA to show the performance ceiling if the model can benefit from learning all datasets at once. The left and right parts of the table shows the balance between stability and plasticity. Overall, our centralized model manages to improve the error rate by $24.6\%$ relatively, compared to the knowledge-base Whisper model, with a $11.7\%$ improvement in backward transfer. Without any German or Arabic data, regularization with Stochastic weight averaging and distillation also avoids catastrophic forgetting, but having more modest improvement compared to our approach. 

The ceiling (Row 2) shows that there is still an undesirable performance loss from averaging the adapters. Both of the continual learning models only achieve about $58\%$ of the maximum possible improvement achieved in the ideal scenario (when the model has full access to all of the datasets, and there is no requirement for backward stability). For these 6 datasets, our model is noticeably better than SWADT in Fisher, TalCS and Ascend (about $10\%$ relatively lower in error rate, and being comparable in ArZen, TunSwitch and SEAME.).

\section{Conclusion}
In this paper, we proposed a continual learning strategy motivated from the wake-sleep cycle of human brain activity, realized by having two separated phases of learning via factorization and consolidation via centralization. The main idea is to separate the input data into segments and maintain different model extensions, followed by a centralization stage merging those extensions back into the knowledge base. We empirically found that using low-rank adapters for factorization and simple averaging the accumulated adapters over time can effectively mitigate catastrophic forgetting while remaining a reasonable performance in learning from the new data, compared to a competitive continual learning approach using weight averaging and distillation. Further works involve expanding the experimental scales with larger datasets, but more importantly explaining the effectiveness of using such simple intuition. Also, our approach needs further modification to be used in online continual learning, in which the model needs to be instantly serviceable given new datasets, or even new samples.

\section{Acknowledgements}
This work was supported by BMBF (grant 01EF1803B), the EU Horizon program (grant 101135798, Meetween), the Helmholtz Association, the HoreKa supercomputer funded by the Ministry of Science, Research and the Arts Baden-Württemberg and BMBF, and Zoom Video Communications grant.


\bibliographystyle{IEEEtran}
\bibliography{mybib}

\end{document}